# Calculation of Entailed Rank Constraints in Partially Non-Linear and Cyclic Models

Peter Spirtes
Department of Philosophy
Carnegie Mellon University
ps7z@andrew.cmu.edu

**Abstract**

The Trek Separation Theorem (Sullivant et al. 2010) states necessary and sufficient conditions for a linear directed acyclic graphical model to entail for all possible values of its linear coefficients that the rank of various sub-matrices of the covariance matrix is less than or equal to $n$, for any given $n$. In this paper, I extend the Trek Separation Theorem in two ways: I prove that the same necessary and sufficient conditions apply even when the generating model is partially non-linear and contains some cycles. This justifies application of constraint-based causal search algorithms to data generated by a wider class of causal models that may contain non-linear and cyclic relations among the latent variables.

## 1 INTRODUCTION

In many cases, scientists are interested in inferring causal relations between variables that cannot be directly measured (e.g. intelligence, anxiety, or impulsiveness) by administering test surveys with measured "indicators" that indirectly measure the unmeasured or "latent" variables. In other cases, scientists are interested in estimating the values of the latent variables from the measured indicators. The variances of the estimates of the latent variables of interest can be reduced in various ways by employing multiple indicators for each latent variable. A model in which each latent variable of interest is measured by multiple indicators (which may also be caused by other latents of interest as well as the error variable) is called a multiple indicator model. Multiple indictor models are quite common in many disciplines such as educational research, psychology, political science, etc. (Bartholomew et al., 2002).

Two major problems are how to use the values of the measured indicator variables to make reliable inferences about the causal relationships between the latent variables of interest, and to predict the values of the latent variables from the values of the measured indicators. A number of complications make both of these tasks very difficult:

- Associations among indicators are often confounded by additional unknown latent common causes;
- One indicator may directly affect other indicators (e.g. "anchoring effects");
- There are often a plethora of alternative causal models that are consistent with the data and with the prior knowledge of domain experts, often far too many models to test individually;
- There may be non-linear dependencies among latent variables;
- There may be feedback relationships among latent variables.

The most common algorithms for using measured indicators to find causal relations among latent variables or to infer the values of the latent variables use some version of factor analysis. However, given the models with the features cited above, factor analytic algorithms, as well as the FindHidden algorithm of Elidan (2001) have been shown to perform poorly (Silva et al. 2006).

One class of model search algorithms that have had some success dealing with some of the complications listed above is constraint-based search. A constraint-based search attempts to find the set of models that most closely match the measured constraints on a probability distribution that are entailed for all values of the free parameters (e.g. conditional independence constraints that are entailed by d-separation) with constraints that are judged to hold in the population (as determined by a statistical test).

Although multiple indicator models rarely entail any conditional independence constraints among just the measured indicators, multiple indicator models often entail constraints on the rank of sub-matrices of the covariance matrix among the measured indicators (e.g. vanishing tetrad differences explained below), and there are searches based on these rank constraints that have desirable properties (Silva et al. 2006).

Multiple indicator models are special cases of structural equation models, and the form of the equations can be represented by a directed graph (Pearl 2000, Spirtes et al. 2001). Under the assumption of linearity, the graphical structure representing the multiple indicator model can *linearly entail* constraints on the covariance matrix of the variables, that is, constraints that hold for all values of the free parameters (the linear coefficients associated with the edges, and the variances of the error terms). For example, a multiple indicator model represented by a graph with a single latent variable $L$ that is the parent of measured indicators $X$, $Y$, $Z$, and $W$ and contains no other edges, entails the vanishing tetrad difference (e.g. $\rho(X,Y)\rho(Z,W) - \rho(X,Z)\rho(Y,W) = 0$) for all values of the linear coefficients, which is equivalent to a constraint that the rank of a submatrix of the covariance matrix is less than or equal to 1.

The Trek Separation Theorem (Sullivant et al. 2010) states necessary and sufficient conditions for a directed acyclic graph to linearly entail that the rank of various sub-matrices of the covariance matrix among the measured variables are less than or equal to $n$, for any $n$.

The Trek Separation Theorem is one way to justify the correctness of the BuildPureClusters algorithm (Silva et al. 2006), that searches for the set of multiple indicator models that most closely match the set of vanishing tetrad differences judged to hold in the population by application of statistical tests to the sample data. BuildPureClusters is a pointwise consistent algorithm that, depending upon the input data, either outputs "Can't tell" or an equivalence class of graphs that linearly entail the same set of vanishing tetrad differences and zero partial correlation constraints. The algorithm has been successfully applied to a number of data sets (Silva et al. 2006, Jackson & Scheines 2005)

However, there are a number of significant limitations on usefulness of the Trek Separation Theorem (and hence on the BuildPureClusters algorithm):

- The Trek Separation Theorem does not apply to cyclic graphs (as in feedback models);
- The Trek Separation Theorem does not apply if any of the causal relations between the variables are non-linear.

In this paper, I prove an extension of the trek separation theorem which gives necessary and sufficient conditions for a directed graph (cyclic or acyclic) that has some functions relating variables to other variables that are non-linear, and in which there may be some feedback (represented by cyclic graphs) to entail that the rank of various sub-matrices of the covariance matrix are less than or equal to $n$, for any $n$. This theorem has at least two uses for causal discovery: it serves as the basis for proving that existing algorithms for the linear case can be reliably applied to partially non-linear or cyclic models (described in section 4), and it could be used in the development of new algorithms for causal inference among models in which measured indicators have multiple latent parents but have non-linear or cyclic relations among the latent parents.

In section 2, I describe multiple indicator models and the Trek Separation Theorem in more detail. In section 3, I state an extension of the trek separation theorem that applies to graphs that may have cyclic and non-linear relationships among some variables. In section 4, I discuss the issue of the extent to which it is to be expected that rank constraints on the covariance matrix might hold, or approximately hold, in the population even if they are not entailed by the model to hold for all values of the free parameters of the model. In section 5, I describe open research questions. The Appendix contains the proofs.

## 2 STRUCTURAL EQUATION MODELS

In what follows, random variables are in italics, and sets of random variables are in boldface..

In a structural equation model (SEM) the random variables are divided into two disjoint sets, the substantive variables (typically the variables of interest) and the error variables (summarizing all other variables that have a causal influence on the substantive variables) (Bollen, 1989). Corresponding to each substantive random variable $V$ is a unique error term $\varepsilon_V$. A *fixed parameter SEM S* has two parts $<\phi, \theta>$, where $\phi$ is a set of equations in which each substantive random variable $V$ is written as a function of other substantive random variables and a unique error variable, together with $\theta$, the joint distributions over the error variables. An example of a linear SEM is the case where $\phi$ contains the pair of linear equations $X = 3L + \varepsilon_X$, and $L = \varepsilon_L$, and $\theta$ is a standardized Gaussian distribution over $\varepsilon_X$ and $\varepsilon_L$ and $\varepsilon_X$ and $\varepsilon_L$ are independent. Together $\phi$ and $\theta$ determine a joint distribution over the substantive variables in $S$, which will be referred to as the *distribution entailed by S*.

A *free parameter* linear SEM model replaces some of the real numbers in the equations in $\phi$ with real-valued variables and a set of possible values for those variables, e.g. $X = a_{X,L} L + \varepsilon_X$, where $a_{X,L}$ can take on any real value. In addition, a free parameter SEM can replace the particular distribution over $\varepsilon_X$ and $\varepsilon_L$ with a parametric family of distributions, e.g. the bi-variate Gaussian distributions with zero covariance. The free parameter SEM also has two parts $<\Phi, \Theta>$, where $\Phi$ contains the set of equations with free parameters and the set of values the free parameters are allowed to take, and $\Theta$ is a family of distributions over the error variables.

In general, I will assume that there is a finite set of free parameters, and all allowed values of the free parameters lead to fixed parameter SEMs that have a reduced form

(i.e. each substantive variable *X* can be expressed as a function of the error variables of *X* and the error variables of its ancestors), all variances and partial variances among the substantive variables are finite and positive, and there are no deterministic relations among the measured variables.

The path diagram of a SEM with jointly independent errors is a directed graph, written with the conventions that it contains an edge $B \to A$ if and only if *B* is a non-trivial argument of the equation for *A*. The error variables are not included in the path diagram. A fixed-parameter acyclic structural equation model (without double-headed arrows) is an instance of a Bayesian Network $<G, P(\mathbf{V})>$, where the path diagram is *G*, and $P(\mathbf{V})$ is the joint distribution over the variables in *G* entailed by the set of equations and the joint distribution over the error variables (Pearl, 2000; Spirtes et al. 2001). It has been shown that when a directed cyclic graph is used to represent non-linear structural equations, then d-separation between **A** and **B** conditional on **C** does not entail the corresponding conditional independence. Even in non-linear cyclic structural equation models, if **A** and **B** are d-separated conditional on the empty set, then **A** and **B** are entailed to be independent (Spirtes, 1995), and that is the only feature of cyclic graphs that the proofs below depend upon.

A polynomial equation *Q* on the entries of a covariance (or correlation) matrix **C** *holds* when **C** is a solution to *Q*. A polynomial *Q* is *entailed by a free parameter* SEM when all values of the free parameters entail covariance matrices that are solutions to *Q*.

For example, a *vanishing tetrad difference* among $\{X,W\}$ and $\{Y,Z\}$, which holds if $\rho(X,Y)\rho(Z,W) - \rho(X,Z)\rho(Y,W) = 0$, is entailed by a free parameter SEM *S* in which *X*, *Y*, *Z*, and *W* are all children of just one latent variable *L* since any value of the free parameters in *S* entails a covariance matrix that is a solution to $\rho(X,Y)\rho(Z,W) - \rho(X,Z)\rho(Y,W) = 0$.

The following definitions are illustrated in Figure 1. A *trek* in G from *I* to *J* is an ordered pair of directed paths $(P_1; P_2)$ where $P_1$ has sink *I*, $P_2$ has sink *J*, and both $P_1$ and $P_2$ have the same source *k* (e.g. $(<L_1,X_1>;<L_1,X_2>)$. The common source *k* is called the *top* of the trek, denoted $top(P_1; P_2)$ (e.g. $top(<L_1,X_1>;<L_1,X_2>)$ is $L_1$). Note that one or both of $P_1$ and $P_2$ may consist of a single vertex, i.e., a path with no edges. A trek $(P_1; P_2)$ is *simple* if the only common vertex between $P_1$ and $P_2$ is the common source $top(P_1; P_2)$. Let **A**, **B**, be two disjoint subsets of vertices **V** in *G*. Let **T(A,B)** and **S(A,B)** denote the sets of all treks and all simple treks from a member of **A** to a member of **B**, respectively. For example, if $\mathbf{A} = \{X_1\}$ and $\mathbf{B} = \{X_2\}$, $\mathbf{S(A,B)} = \{(<L_1,X_1>;<L_1,X_2>); (<L_2,X_1>;<L_2,X_2>)\}$.

For two sets of variables **A** and **B**, and a covariance or correlation matrix over a set of variables **V** containing **A** and **B**, let cov(**A**, **B**) be the sub-matrix of **Σ** that contains the rows in **A** and columns in **B**. For example, if $\mathbf{A} = \{X_1, X_2, X_3\}$, and $\mathbf{B} = \{X_4, X_5, X_{10}\}$, then

$$\text{cov}(\mathbf{A},\mathbf{B}) = \begin{array}{c} \\ X_1 \\ X_2 \\ X_3 \end{array} \begin{array}{c} X_4 \quad\quad X_5 \quad\quad X_{10} \\ \left[\begin{array}{ccc} \rho(X_1,X_4) & \rho(X_1,X_5) & \rho(X_1,X_{10}) \\ \rho(X_2,X_4) & \rho(X_2,X_5) & \rho(X_2,X_{10}) \\ \rho(X_3,X_4) & \rho(X_3,X_5) & \rho(X_3,X_{10}) \end{array}\right] \end{array}$$

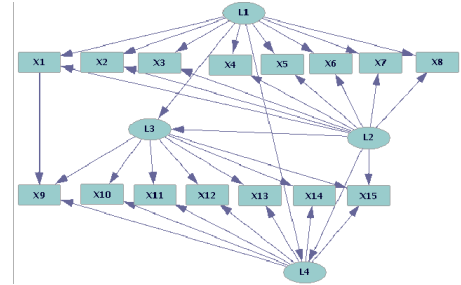

**Figure 1: A Multiple Indicator Model**

In the case where **A** and **B** both have size 3, if the rank of the matrix is less than or equal to 2, the determinant of cov(**A**,**B**) = 0. In that case the matrix is said to satisfy a *sextad constraint*. An example of a sextad constraint is

$$Det\left(\left[\begin{array}{ccc} \rho(X_1,X_4) & \rho(X_1,X_5) & \rho(X_1,X_{10}) \\ \rho(X_2,X_4) & \rho(X_2,X_5) & \rho(X_2,X_{10}) \\ \rho(X_3,X_4) & \rho(X_3,X_5) & \rho(X_3,X_{10}) \end{array}\right]\right) = 0$$

Let **A**, **B**, $\mathbf{C_A}$, and $\mathbf{C_B}$ be four subsets of the set **V** of vertices in *G*, which need not be disjoint. The pair ($\mathbf{C_A}$;$\mathbf{C_B}$) *trek separates* (or *t-separates*) **A** from **B** if for every trek ($P_1$; $P_2$) from a vertex in **A** to a vertex in **B**, either $P_1$ contains a vertex in $\mathbf{C_A}$ or $P_2$ contains a vertex in $\mathbf{C_B}$; $\mathbf{C_A}$ and $\mathbf{C_B}$ are *choke sets* for **A** and **B**. For example, ($\{L_1\}$; $\{L_2\}$), ($\{L_1, L_2\}$; $\varnothing$), and ($\varnothing$; $\{L_1,L_2\}$) all t-separate **A** from **B** in this example.

**Theorem 1 (Trek Separation Theorem):** For all directed acyclic graphs (path diagrams) *G*, the sub-matrix cov(**A**,**B**) has rank less than or equal to *r* for all covariance matrices of linear SEMs with path diagram *G*, if and only if there exist subsets $\mathbf{C_A}$, $\mathbf{C_B} \subset \mathbf{V}(G)$ with $\#\mathbf{C_A} + \#\mathbf{C_B} \leq r$ such that ($\mathbf{C_A}$; $\mathbf{C_B}$) t-separates **A** from **B** (where $\#\mathbf{C_A}$ is the number of variables in $\mathbf{C_A}$, and $\mathbf{V}(G)$ is the set of vertices in *G*). (Sullivant et al., 2010)

Since the rank of cov(**A**, **B**) is less than or equal to *r*, if $\mathbf{C_A} \cap \mathbf{C_B} = \varnothing$, $\#\mathbf{A} = \#\mathbf{B} = 3$, $\#\mathbf{C_A} + \#\mathbf{C_B} = 2$, and ($\mathbf{C_A}$; $\mathbf{C_B}$) t-separates **A** from **B**, then *G* entails a sextad constraint among the variables in **A** and **B**. For example, in Figure 1, ($\{L_1, L_2\}$;{ }) trek separates $\{X_1, X_2, X_3\}$ from $\{X_4, X_5, X_{10}\}$, and hence

$$\text{rank}\left(\left[\begin{array}{ccc} \rho(X_1,X_4) & \rho(X_1,X_5) & \rho(X_1,X_{10}) \\ \rho(X_2,X_4) & \rho(X_2,X_5) & \rho(X_2,X_{10}) \\ \rho(X_3,X_4) & \rho(X_3,X_5) & \rho(X_3,X_{10}) \end{array}\right]\right) \leq \#\mathbf{C_A} + \#\mathbf{C_B} = 2$$

which in turn entails that the determinant of the matrix is zero for all values of the free parameters in a linear SEM.

## 3 AN EXTENSION OF THE TREK SEPARATION THEOREM

The Trek Separation Theorem can be extended by weakening the assumptions that the graph be linear everywhere and acyclic everywhere. The exact definition of *linear acyclicity* (or LA for short) *below* a *choke* set is somewhat complex (and is given below), but roughly a directed graphical model is *LA below sets* ($C_A$; $C_B$) for $A$ and $B$ respectively, if there are no directed cycles between $C_A$ and $A$ or $C_B$ and $B$, and for every vertex $V$ on any directed path $P$ from $C_A$ to $A$, $V$ is a linear function of its parents along $P$ plus an arbitrary function of the parents not along $P$ (including the error variables); and similarly for $C_B$ and $B$. For example in Figure 1, let the sets $C_A$ and $C_B$ for $A = \{X_1, X_2, X_3\}$, and $B = \{X_4, X_5, X_{10}\}$ be $C_A = \{L_1, L_2\}$ and $C_B = \varnothing$. Linear acyclicity below the sets $C_A$, $C_B$, for $A$, $B$ requires that for $i = 1\ldots 3$, $X_i = a_{i,1} L_1 + a_{i,2} L_2 + f_i(\varepsilon_i)$, where $\varepsilon_i$ is the error term for $X_i$, and $f_i$ is an arbitrary measurable function. (Since $C_B = \varnothing$, linear acyclicity below the set $C_B$ is trivially true). Note that there can be non-linear and/or cyclic relationships between any of the latent variables.

More formally, let $D(C_A, A, G)$ be the set of vertices on directed paths in $G$ from $C_A$ to $A$ except for the members of $C_A$ (but including members of $A \setminus C_A$). If $S$ is a fixed-parameter SEM $<\phi, \theta>$ with path diagram $G$, $S$ is LA *below the sets* $C_A$, $C_B$ *for* $A$, $B$ iff for each member of $W = D(C_A, A, G) \cup D(C_B, B, G)$,

(i) $V_{ext} = V \cup \{\varepsilon_X : X \in W\}$;

(ii) no member of $W$ lies on a cycle;

(iii) $G_{ext}$ is a directed graph over $V_{ext}$ with sub-graph $G$, together with an edge from $\varepsilon_X$ to $X$ for each $X \in W$,

(iv) for each $X \in D(C_A, A, G_{ext})$,

$$X = \sum_{V \in \mathbf{Parents}(X, G_{ext}) \cap (D(C_A, A, G_{ext}) \cup C_A)} a_{X,V} V + f_X(\mathbf{Parents}(X, G_{ext}) \setminus (D(C_A, A, G_{ext}) \cup C_A)) \quad (1)$$

and for each $X \in D(C_B, B, G_{ext})$,

$$X = \sum_{V \in \mathbf{Parents}(X, G_{ext}) \cap (D(C_B, B, G_{ext}) \cup C_B)} a_{X,V} V + g_X(\mathbf{Parents}(X, G_{ext}) \setminus (D(C_B, B, G_{ext}) \cup C_B)) \quad (2)$$

Note that $D(C_A, A, G) = D(C_A, A, G_{ext})$ for any $A$ and $C_A$ that do not contain an error variable.

**Theorem 2 (Extended Trek Separation Theorem):** Suppose $G$ is a directed graph containing $C_A$, $A$, $C_B$, and $B$, and $(C_A; C_B)$ t-separates $A$ and $B$ in $G$. Then for all covariance matrices entailed by a fixed parameter structural equation model $S$ with path diagram $G$ that is LA below the sets $C_A$ and $C_B$ for $A$ and $B$, $rank(cov(A,B)) \leq \#C_A + \#C_B$.

The converse of Theorem 2 is basically guaranteed by the "only-if" clause of Theorem 1.

**Theorem 3:** For all directed graphs $G$, if there does not exist a pair of sets $C'_A$, $C'_B$, such that $(C'_A; C'_B)$ t-separates $A$ and $B$ and $\#C'_A + \#C'_B \leq r$, then for any $C_A$, $C_B$ there is a fixed parameter structural equation model $S$ with path diagram $G$ that is LA below the sets $(C_A; C_B)$ for $A$ and $B$ that entails $rank(cov(A,B)) > r$.

In order to use the Extended Trek Separation theorems, it is necessary to have statistical tests of when rank constraints hold, or equivalently, when the corresponding determinants are zero. Drton & Olkin (2008) describe a statistical test of the rank constraints, that assumes a Normal distribution; however, in practice even when the distributions is non-Normal, the test often performs well. The Wishart test for vanishing tetrad constraints is a special case of this test (and was used in all of the simulations performed.)

There is also a much slower, but asymptotically distribution-free statistical test of rank constraints based on the test developed by Bollen and Ting (Bollen & Ting, 1993).

## 4 FAITHFULNESS

Let a distribution $P$ be *linearly rank-faithful* to a directed acyclic graph $G$ if every rank-constraint on a sub-covariance matrix that holds in $P$ is entailed by every free-parameter linear structural equation model with path diagram equal to $G$.

If a distribution is linearly rank-faithful to its causal graph, then it is possible to use the rank-constraints among the observed variables to draw conclusions about the t-separation structure of the causal graph by using the Trek Separation Theorem to identify latent choke sets. For example, given a quartet of variables $V = \{X_1, X_2, X_3, X_4\}$, if for every partition of $V$ into two sets of equal size (e.g. $A = \{X_1, X_2\}$, $B = \{X_3, X_4\}$) the rank of cov($A,B$) is 1, this indicates that there are sets $C_A$ with one member and $C_B = \varnothing$ such that $(C_A; C_B)$ t-separates($A; B$). By combining this with other rank constraints and partial correlation constraints, it is possible to conclude, e.g. that $X_1$, $X_2$, $X_3$ and $X_4$ have a single latent common cause (Silva et al. 2006)

In practice, there is no oracle that states whether a given rank constraint holds in a population, so statistical tests of rank constraints are substituted for an oracle. But is the assumption of linear rank-faithfulness reasonable? One justification for the assumption of rank-faithfulness is that the Trek Separation Theorem entails that if there is no pair of sets $C_A$ and $C_B$ such that $\# C_A + \# C_A \leq r$, and $A$ and $B$ are t-separated by $(C_A; C_B)$ then the rank of cov($A,B$) is not linearly entailed to be $\leq r$ for all values of the free parameters of a free parameter structural equation model with path diagram *G*. Moreover, since cov(**A**,**B**) is a linear function of the covariance matrix among the latents and the covariance matrix of the error terms, and the rank is not linearly entailed to be of rank *r* or less, it follows that the set of values of free parameters for which rank(cov(**A**,**B**)) $\leq r$ is of Lebesgue measure 0. This fact can be used to demonstrate the pointwise consistency of algorithms that rely on statistical tests of rank-constraints (Silva et al. 2006) under the assumption of linear rank-faithfulness.

This does not settle the practicality of such algorithms on reasonable sample sizes. Since statistical tests of the rank constraints are used to determine whether or not a rank-constraint holds in a population, if the relevant determinants that determine rank are very close to, but not exactly equal to zero, any algorithm relying on statistical tests of rank could be incorrect with high probability unless the sample sizes were unrealistically large. This can occur for example, when some of the correlations between observed indicators are either very close to zero or very close to 1. Nevertheless, simulation tests and real applications are positive evidence that BuildPureClusters works at reasonable sample sizes. For a further discussion of faithfulness assumptions see Spirtes et al. (2001), Robins et al. (2003), Kalisch & Buhlmann (2007), and Uhler et al. (2012).

The concept of linear rank faithfulness can be extended in the following way. If $\Phi$ is a set of functions that contains the linear functions as a special case, a distribution *P* is <$\Phi$, $\Theta$>-*LA below the sets* **C$_A$**, **C$_B$** *for* **A**, **B** *rank faithful* to a directed graph *G* if every rank constraint that holds in *P* is entailed to hold by every free parameter SEM <$\Phi$, $\Theta$> with path diagram *G* that is LA below the sets **C$_A$**, **C$_B$** *for* **A**, **B**.

Suppose in what follows that a given free parameter structural equation model *S* = <$\Phi$, $\Theta$> is LA below the sets (**C$_A$**; **C$_B$**) for **A**, **B**, and that for each equation *X* = *f*(**Y**) in $\Phi$ not required to be linear by definition, a linear equation with any value of the coefficients

$$X = \sum_{Y \in \mathbf{Y}} a_{X,Y} Y$$

is the result of a substitution of some value for the free parameters in *S*. For example, if $X = a_1Y + a_2Y^2$, then for any value of $a_1$, $X = a_1Y$ is the result of setting the free parameter $a_2$ to zero. In contrast, if $\Phi$ contained only $X = a_2Y^2$, the correlations between *X* and *Y* would be forced to be zero for all $a_2$, which in general could lead to rank constraints holding for all values of the free parameters even without the corresponding t-separation relations holding in *G*.

If all of the variables in $\Phi$ are analytic functions, whenever the set of solutions to an analytic function is not the entire space of values, the set of solutions has Lebesgue measure 0 (Kilmer et al. 1996). So the same kind of argument for faithfulness in the LA-below-the-choke-set case can be made as in the linear case, as long as $\Phi$ contains all LA functions among the part of the graph that is not below the choke sets as a special case.

This still leaves the question of whether there are common "almost" violations of rank faithfulness that could only be discovered with enormous sample sizes (i.e. the relevant determinants are very close to zero).

In order to illustrate one use of the extension of the Trek Separation Theorem and to do a preliminary test of the extent to which the introduction of non-linearity makes the problem of almost violations of the assumption of rank-faithfulness more common, I performed a simulation study of the Silva et al. BuildPureClusters Algorithm, using both linear models, and LA-below the choke set models.

The BuildPureClusters Algorithm (Silva et al. 2006) takes as input sample data and attempts to find a subset **S** of the measured indicators such no two members of **S** have a directed edge between them, no member of **S** has more than one latent parent, and the measured variables in **S** are partitioned into clusters, where each member of a cluster is the child of the same latent parent. (This is useful for determining which measured variables are measuring which latent variables, and is input to the MIMBuild algorithm that searches for the causal structure among the latent variables.) BuildPureClusters uses tests of vanishing tetrad differences to select and cluster the variables (which are equivalent to tests of whether various $2 \times 2$ submatrices of the covariance matrix have rank 1.) Not all of the rank tests that BuildPureClusters uses in general are also entailed for the case where the relationships between the latents are non-linear (which is *not* the same as LA below the choke sets), but all of the ones that it uses for this particular study are entailed in the non-linear case.

Figure 2 illustrates a model that contains an impure measurement model because of the $X_1 \rightarrow X_6$ edge and because $X_{10}$ has two latent parents (indicated by the red arrows) while Figure 3 illustrates that if $X_6$ and $X_{10}$ are removed, the resulting model has a pure measurement model. Thus correct output for Figure 2 would either be $\{X_1, X_2, X_3, X_4, X_5\}$ and $\{X_7, X_8, X_9\}$ or $\{X_2, X_3, X_4, X_5\}$ and $\{X_6, X_7, X_8, X_9\}$.

The model in the simulation contained 5 latent variables ($L_1$ through $L_5$), each with 5 measured children ($X_1$ through $X_{25}$), with $L_2$ through $L_5$ children of $L_1$. It also contained edges $X_1 \rightarrow X_6$, $X_{15} \rightarrow X_{19}$, $L_3 \rightarrow X_{10}$, and $L_4 \rightarrow X_{21}$, which introduced impurities. The input to the algorithm in each case was raw data at one of 3 sample sizes, 100, 500, and 1000. Each variable is a linear function of its parents plus a unique error term, where the linear coefficients were chosen uniformly from the range 0.5 to 2.0, and the error terms were independent standard Gaussian. Each latent variable $L_i$ ($i = 2…5$) was equal to $aL_1 + bcL_1^3 + \varepsilon_i$, where *a* was chosen uniformly from 0.25 to 1.0, *c* was chosen uniformly from 0.5 to 2.0, and the degree of non-linearity was varied by setting *b* to each of the values 0.0, 0.01, 0.02, 0.03, and 0.05 in turn. The degree of non-linearity of the relationship between the measured variables was measured by the median p-value of the White test of non-linearity between each pair of measured variables (which is 0.5 for linear relationships).

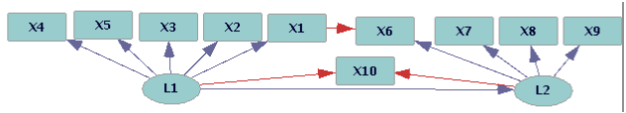

**Figure 2:** Impure Measurement Model

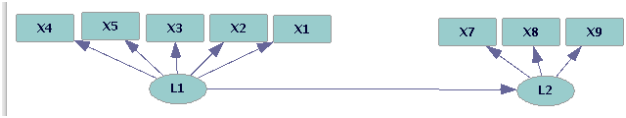

**Figure 3:** Pure Measurement Submodel

In order to avoid detectable cases of almost unfaithful rank constraints, if the correlation matrix of the observed indicators contained correlations close to zero (less than 0.09) or close to 1 (greater than .9) the data was rejected. (In actual practice, instead of rejecting the data a user could simply search for a subset of variables that did not contain extreme correlations.) The simulation set the p-value used in the algorithm to 0.01 for every case, and the TETRAD IV implementation was used (http://www.phil.cmu.edu/projects/tetrad/current.html).

The correctness of the output of BuildPureClusters was measured in three ways:

1. How many clusters the algorithm found (which was a maximum of 5 in each case).
2. How far a given output cluster was from being a pure cluster. I set Purity for a given output cluster to Purity = (Size of largest pure subcluster contained in output cluster)/(size of output cluster). For example, if the output cluster for data generated by the a model had 7 variables {$X_1$, $X_2$, $X_3$, $X_4$, $X_5$, $X_6$, $X_9$}, and $X_1 - X_6$ were all children of latent variable $L_1$, and $X_9$ was a child of latent variable $L_2$, $X_9$ would have to be removed in order to make the output cluster pure (leaving 6 variables), so Purity for the output cluster would be equal to 6/7.
3. The percentage of the largest pure actual subclusters included in the output. I set Fraction size = (Size of the output cluster)/(size of the largest actual pure subcluster containing it). For example, if a model has an actual pure subcluster of size 8 (e.g. $X_1 - X_8$) and if the output contained a corresponding subcluster of size 6 (e.g. $X_1 - X_6$) then Fraction size for the output cluster is 6/8. (If the output contained only four subclusters instead of the potential five subclusters, I calculated this only for the subclusters that were 100 data sets were generated at each sample size, for both the linear and non-linear case. Then the BuildPureClusters algorithm was applied to each data set, using the Wishart test of vanishing tetrad differences. The Wishart test assumes the variables have joint Gaussian distributions, which is true in the linear Gaussian case but not the non-linear case.

| Size | Cubic | Cluster Number | Average Purity | Average Fraction | Median White |
|------|-------|----------------|----------------|------------------|--------------|
| 100  | 0.00  | 3.89 | .909 | .782 | .500 |
| 100  | 0.01  | 4.26 | .930 | .792 | .414 |
| 100  | 0.02  | 4.32 | .931 | .806 | .291 |
| 100  | 0.03  | 4.26 | .935 | .809 | .285 |
| 100  | 0.05  | 4.29 | .937 | .809 | .241 |
| 500  | 0.00  | 4.34 | .957 | .820 | .508 |
| 500  | 0.01  | 4.41 | .953 | .813 | .349 |
| 500  | 0.02  | 4.34 | .950 | .813 | .119 |
| 500  | 0.03  | 4.29 | .954 | .813 | .088 |
| 500  | 0.05  | 4.48 | .957 | .829 | .0001 |
| 1000 | 0.00  | 4.78 | .930 | .900 | .510 |
| 1000 | 0.01  | 4.91 | .953 | .926 | .288 |
| 1000 | 0.02  | 4.55 | .924 | .909 | .030 |
| 1000 | 0.03  | 4.31 | .912 | .899 | .017 |
| 1000 | 0.05  | 4.52 | .956 | .831 | $4.43e^{-10}$ |

**Table 1:** Output of First Simulation Study

The results of the simulation test are summarized in Table 1, where each row gives values for 100 runs of a given kind. The columns in order give the sample size, the value of the *b* coefficient, the average number of clusters, average Purity, average Fraction size, and median p-value of a White test of non-linearity applied to each pair of measured variables. The maximum correct number of clusters is 5, the maximum average Purity is 1, and the maximum average Fraction is 1.

The results of the simulation study (Table 1) indicate that at least with respect to vanishing tetrad differences, the BuildPureClusters algorithm performs about as well in the nonlinear and the linear case using the Wishart test. There is no systematic advantage of linear over non-linear or vice-versa, and the results are generally close in both cases. Hence, in this limited test, the non-linearity that was introduced did not make the problem of almost unfaithful rank constraints much worse in terms of the output.

A simulation study of the extent to which violation of the assumption that the observed variables are linearly related to their latent parents affects the performance of the BuildPureClusters algorithm was also performed. The input to the algorithm in each case was raw data at one of 2 sample sizes, 100 and 1000. The latent variables were simulated in the same was as in the previously described simulation. Each measured variable was set equal to $(1 - d)eL_i + dfL_i^3 + \varepsilon_i$ (where $L_i$ is the parent of the measured variable in the graph), *e* and *f* were independently selected from a uniform distribution between 0.5 and 2.0, and the degree of non-linearity of the relationship between the measured and the latents was varied by setting *d* to either 0.01 or 0.05 in turn. The error terms were independent standard Gaussians. The results are shown in Table 2, where the second column reports the values both of *b* (the first number, from the equation for the relationships between the latents) and *d* (the second number, from the equations relating the measured variables to their latent parent.)

As with the previously described simulation, the result of making the relationships between the latents non-linear does not have any systematic effect on the performance of the BuildPureClusters Algorithm However, as the nonlinearity of the relationship between the measured variables and their latent parents increases, the output becomes much less informative (as evidenced by the large decreases in the Number of Clusters, and the Average Fraction), but is generally not incorrect (as evidenced by the small decreases in the Average Purity). When the assumption of linear relationships between the measured and latent variables is violated, the algorithm actually performs better at smaller sample sizes, presumably because at larger sample sizes even small deviations from the assumption lead to rejection of the rank constraints.

| Sample Size | Cubic | Number of Clusters | Average Purity | Average Fraction | Median White |
|---|---|---|---|---|---|
| 100 | 0:.01 | 3.42 | .909 | .755 | .302 |
| 100 | 0:.05 | 2.65 | .874 | .668 | .205 |
| 100 | .05:.01 | 2.65 | .903 | .754 | .346 |
| 100 | .05:.05 | 3.23 | .902 | .679 | .212 |
| 1000 | 0:.01 | 2.21 | .942 | .713 | .019 |
| 1000 | 0:.05 | 0.72 | .868 | .344 | 6.1e$^{-4}$ |
| 1000 | .05:.01 | 3.22 | .942 | .749 | .106 |
| 1000 | .05:.05 | 1.20 | .895 | .305 | 6.9e$^{-4}$ |

**Table 2:** Output of Second Simulation Study

## 5 OPEN QUESTIONS

In this paper I have proved that the necessary and sufficient conditions for a class of rank constraints on submatrices of a covariance matrix to be implied by a linear model can be extended to models that contains some non-linear and/or cyclic relationships. This shows that existing algorithms that use these rank constraints to search for causal models can be reliably applied to a much wider class of models, as long a rank-faithfulness condition holds. I also argued that the same kind of considerations that argue for rank-faithfulness in linear models can be extended to some kinds of non-linear structure equation models.

In order to make full use of this theorem, it would be very helpful to have a computationally feasible test of when two models are equivalent with respect to rank constraints of a given kind assuming they are both LA below their choke sets. Nor is it known how to graphically represent the features that each member of such an equivalence class has in common. In addition, the question of the extent to which almost violations of faithfulness are made worse by different classes of non-linear functional relationships among variables also needs to be more fully investigated.

## 6 APPENDIX

The proof of Theorem 2 requires the following two lemmas.

**Lemma 1:** Suppose that $\mathbf{C_A} \neq \emptyset$, $\mathbf{A} = \mathbf{\Lambda_A C_A} + \mathbf{f(E_A)}$, and $\text{cov}(\mathbf{f(E_A)},\mathbf{B}) = 0$, where $\mathbf{\Lambda_A}$ is a #$\mathbf{A}$ by #$\mathbf{C_A}$ matrix of real numbers. Then $rank(\text{cov}(\mathbf{A},\mathbf{B})) \leq$ #$\mathbf{C_A}$.

**Proof.**

$\text{cov}(\mathbf{A},\mathbf{B}) = \text{cov}(\mathbf{\Lambda_A C_A} + \mathbf{f(E_A)},\mathbf{B}) = \text{cov}(\mathbf{\Lambda_A C_A},\mathbf{B}) + \text{cov}(\mathbf{f(E_A)},\mathbf{B}) = \mathbf{\Lambda_A}\text{cov}(\mathbf{C_A},\mathbf{B})$. Hence $rank(\text{cov}(\mathbf{A},\mathbf{B}) = rank(\mathbf{\Lambda_A}\text{cov}(\mathbf{C_A},\mathbf{B}))$. It follows that

$rank(\text{cov}(\mathbf{C_A},\mathbf{B})) \leq \min(\#\mathbf{C_A},\#\mathbf{B}) \leq \#\mathbf{C_A}$
$rank(\mathbf{\Lambda_A}) \leq \min(\#\mathbf{C_A},\#\mathbf{A}) \leq \#\mathbf{C_A}$
$rank(\mathbf{\Lambda_A}\text{cov}(\mathbf{C_A},\mathbf{B})) \leq$
$\min(rank(\mathbf{\Lambda_A}),rank(\text{cov}(\mathbf{C_A},\mathbf{B}))) \leq$
$\min(\#\mathbf{C_A},\#\mathbf{C_A}) \leq \#\mathbf{C_A}$

Q.E.D.

Next consider the case where $\mathbf{A}$ is a linear function of $\mathbf{C_A}$ plus a function of a set of variables $\mathbf{E_A}$, $\mathbf{B}$ is a linear function of $\mathbf{C_B}$ plus a function of a set of variables $\mathbf{E_B}$, and all of the variables in $\mathbf{E_A}$ are uncorrelated with all of the variables in $\mathbf{E_B}$.

**Lemma 2:** Suppose that $\mathbf{C_A} \neq \emptyset$, $\mathbf{C_B} \neq \emptyset$, #$\mathbf{C_A} = p$, #$\mathbf{C_B} = q$, #$\mathbf{A} = r$, #$\mathbf{B} = s$, $\mathbf{A} = \mathbf{\Lambda_A C_A} + \mathbf{f(E_A)}$, $\text{cov}(\mathbf{f(E_A)},\mathbf{g(E_B)}) = 0$, $\mathbf{B} = \mathbf{\Lambda_B C_B} + \mathbf{f(E_B)}$. Then $rank(\text{cov}(\mathbf{A},\mathbf{B})) \leq$ #$\mathbf{C_A}$ + #$\mathbf{C_B}$.

**Proof.**

$\text{cov}(\mathbf{A},\mathbf{B}) = \text{cov}(\mathbf{\Lambda_A C_A} + \mathbf{f(E_A)}, \mathbf{\Lambda_B C_B} + \mathbf{g(E_B)}) =$
$\text{cov}(\mathbf{\Lambda_A C_A}, \mathbf{\Lambda_B C_B} + \mathbf{g(E_B)}) + \text{cov}(\mathbf{f(E_A)}, \mathbf{\Lambda_B C_B}) +$
$\text{cov}(\mathbf{f(E_A)}, \mathbf{g(E_B)}) =$
$\mathbf{\Lambda_A}\text{cov}(\mathbf{C_A}, \mathbf{\Lambda_B C_B} + \mathbf{g(E_B)}) + \text{cov}(\mathbf{f(E_A)}, \mathbf{C_B})\mathbf{\Lambda_B}^T$

$rank(\text{cov}(\mathbf{C_A}, \mathbf{\Lambda_B C_B} + \mathbf{g(E_B)})) \leq \min(p,s) \leq p$
$rank(\mathbf{\Lambda_A}) \leq \min(r,p) \leq p$

$rank(\mathbf{\Lambda_A}\text{cov}(\mathbf{C_A}, \mathbf{\Lambda_B C_B} + \mathbf{g(E_B)})) \leq$
$\min(rank(\mathbf{\Lambda_A}), rank(\text{cov}(\mathbf{C_A}, \mathbf{\Lambda_B C_B} + \mathbf{g(E_B)}))) \leq$
$\min(p,p) \leq p$
$rank(\text{cov}(\mathbf{f(E_A)},\mathbf{C_B})) \leq \min(r,q) \leq q$
$rank(\mathbf{\Lambda_B}) \leq \min(q,s) \leq q$
$rank(\text{cov}(f(\mathbf{E_A}),\mathbf{C_B}))\mathbf{\Lambda_B}^T) \leq \min(q,q) \leq q$

It follows that the sum of two matrices of the same number of rows and columns is at most #$C_A$+#$C_B$. Q.E.D.

**Theorem 2:** Suppose $G$ is a directed graph containing $C_A$, **A**, $C_B$, and **B**, and ($C_A$; $C_B$) t-separates **A** and **B** in $G$. Then for all covariance matrices entailed by a fixed parameter structural equation model $S$ with path diagram $G$ that is LA below the sets $C_A$ and $C_B$ for **A** and **B**, $rank(cov(\mathbf{A},\mathbf{B})) \leq$ #$C_A$ + #$C_B$.

**Proof**. In the proof, the graphical relations all refer to $G_{ext}$, so the graphical arguments will be dropped when referring to parents and directed paths. I will prove the theorem by showing that t-separation of **A** and **B** by ($C_A$; $C_B$) entails that **A** can be written as a linear function of $C_A$ plus a function of a set of variables $E_A$, that **B** can be written as a linear function of $C_B$ plus a function of a set of variables $E_B$, and that all of the variables in $E_A$ are uncorrelated with all of the variables in $E_B$. Then applying Lemmas 1 and/or 2 proves the theorem.

Case 1: If $C_A = C_B = \emptyset$, then there are no treks between **A** and **B**. Hence **A** and **B** are jointly independent. It follows that $cov(\mathbf{A},\mathbf{B}) = 0$, which is of rank 0 = #$C_A$+#$C_B$

Case 2: $C_A \neq \emptyset$. I will show that for each $A_i \in \mathbf{A}$,

$$A_i = \sum_{V \in C_A} a_{i,V} V + f_i(\mathbf{E_i})$$

where each member of $\mathbf{E_i}$ is not in $D(C_A, \mathbf{A}) \cup C_A$ and is an ancestor of $A_i$ via some (possibly single-vertex) path that does not intersect $C_A$.

Case 2a: $A_i \in C_A$. Set $A_i = 1 \times A_i$, $\mathbf{E_i} = \emptyset$, and $f_i(\mathbf{E_i}) = 0$. Since $A_i$ is in $C_A$, $A_i$ is a linear function of $C_A$, and trivially each member of $\mathbf{E_i}$ is not in $D(C_A, \mathbf{A}) \cup C_A$ and is an ancestor of $A_i$ via some (possibly single-vertex) path that does not intersect $C_A$.

Case 2b: $A_i \notin C_A$.

Case 2b(i): $D(C_A, A_i) = \emptyset$. Set $\mathbf{E_i} = \{A_i\}$, $f_i(\mathbf{E_i}) = A_i$. By assumption, each member of $\mathbf{E_i}$ is not in $D(C_A, \mathbf{A}) \cup C_A$ and is an ancestor of $A_i$ via some (possibly single-vertex) path that does not intersect $C_A$.

Case 2bii: $D(C_A, A_i) \neq \emptyset$. The longest directed path from $C_A$ to $A_i$ is of finite length. Let $R = \{V \in \mathbf{Parents}(A_i) \cap (D(C_A) \cup C_A)\}$. By the assumption of LA below the choke sets $C_A$, $C_B$, for **A**, **B**,

$$A_i = \sum_R a_{i,V} V + f_i(\mathbf{Parents}(A_i) \setminus (D(C_A, \mathbf{A}) \cup C_A))$$

The algorithms in this section of the proof are illustrated in Figure 4 and Figure 5 (where only the relevant error variables are shown in the graph). For each vertex in $R$, substitute the r.h.s of equation 1 in for $V$. Continue substitutions until no more substitutions based on equation 1 can be made. The proof is by induction on the number of substitutions.

Let $V_i = \mathbf{Parents}(A_i) \cap D(C_A, \mathbf{A}) \cup C_A)$, $f_i^1 = f_i$ and

$\mathbf{E}_i^1 = \mathbf{Parents}(A_i) \setminus (D(C_A, \mathbf{A}) \cup C_A)$ at stage 1 of equation 1. Every member of $\mathbf{E}_i^1$ is not in $D(C_A, \mathbf{A}) \cup C_A$ by definition. An edge from any member of $\mathbf{E}_i^1$ to $A_i$ constitutes a path to $A_i$ that does not intersect $C_A$.

Suppose for an induction hypothesis that after $n$ substitutions,

$$A_i = \sum_{V \in V_n} a_{i,V}^n V + f_i^n(\mathbf{E}_i^n) + a_{i,X}^n X$$

where $V_n \subseteq D(C_A, \mathbf{A}) \cup C_A$, $X \in D(C_A, \mathbf{A}) \cap V_n$, there is no member of $V_n$ whose longest path to $A_i$ is shorter than the longest path from $X$ to $A_i$, and each member of $\mathbf{E}_i^n$ is not a member of $D(C_A, \mathbf{A}) \cup C_A$ but is an ancestor of $A_i$ via a directed path that does not intersect $C_A$. The superscripts represent which substitution the superscripted term first appeared in. If no such $X$ exists (because $A_i$ is expressed as a function of members of $C_A$ and variables that are not on paths from $C_A$ to $\mathbf{A}$), the algorithm is done.

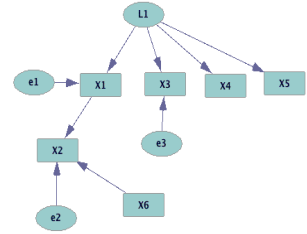

$X_2 = 3 X_1 + f_2(\varepsilon_2, X_6)$   $X_4 = 0.6 L_1 + f_4(\varepsilon_4)$
$X_1 = 2 L_1 + f_1(\varepsilon_1)$   $X_5 = 0.9 L_1 + f_5(\varepsilon_5)$
$X_3 = 0.8 L_1 + f_3(\varepsilon_3)$
$\mathbf{A} = \{X_2, X_3\}$  $\mathbf{B} = \{X_4, X_5\}$  $C_A = \{L_1\}$  $C_B = \emptyset$
$D(C_A, \mathbf{A}) = \{X_1, X_2, X_3\}$  $D(C_B, \mathbf{B}) = \emptyset$

**Figure 4:** Illustration of base stage of substitutions

Otherwise, Let $R = \mathbf{Parents}(X) \cap D(C_A, \mathbf{A}) \cup C_A$. Substitute the r.h.s. of

$$X = \sum_{V \in R} a_{X,V} V + f_X(\mathbf{Parents}(X) \setminus (D(C_A, \mathbf{A}) \cup C_A))$$

in for $X$ in the equation. After the substitution,

$$A_i = \sum_{V \in V_n} a_{i,V}^n V + f_i^n(\mathbf{E}_i^n) +$$

$$a_{i,X}^n \left( \sum_{V \in V_n} a_{X,V} V + f_X(\mathbf{Parents}(X) \setminus (D(C_A, \mathbf{A}) \cup C_A)) \right)$$

$$= \sum_{V \in V_n} a_{i,V}^n V + a_{i,X}^n \left( \sum_{V \in R} a_{X,V} V \right) +$$

$$f_i^n(\mathbf{E}_i^n) + a_{i,X}^n f(\mathbf{Parents}(X) \setminus (D(C_A, \mathbf{A}) \cup C_A)) =$$

$$V_{n+1} = (V_n \cup (\mathbf{Parents}(X) \cap (D(\mathbf{C_A}, \mathbf{A}) \cup \mathbf{C_A}))) \setminus X$$

$$\mathbf{E}_i^{n+1} = \mathbf{E}_i^n \cup (\mathbf{Parents}(X) \setminus (D(\mathbf{C_A}, \mathbf{A}) \cup \mathbf{C_A})),$$

$$a_{i,V}^{n+1} = a_{i,X}^n a_{X,V} \text{ for parents of } X$$

$$f_i^{n+1}(\mathbf{E}_i^{n+1}) = f_i^n(\mathbf{E}_i^n) + a_{i,X} f_X^n(\mathbf{Parents}(X) \setminus (D(\mathbf{C_A}, \mathbf{A}) \cup \mathbf{C_A}))$$

Figure 5 contains an illustration of the substitutions for the example shown in Figure 4.

Each member of $\mathbf{E}_i^{n+1} \cap \mathbf{E}_i^n$ is not on a directed path from $\mathbf{C_A}$ to $\mathbf{A}$ but is an ancestor of $A_i$ via a path that does not intersect $\mathbf{C_A}$ by the induction assumption. $\mathbf{E}_{A'}^{n+1} \setminus \mathbf{E}_{A'}^n \subseteq \mathbf{Parents}(X) \setminus (D(\mathbf{C_A}, \mathbf{A}) \cup \mathbf{C_A})$ and hence not a member of $D(\mathbf{C_A}, \mathbf{A}) \cup \mathbf{C_A}$. Because $X$ is expanded by substitution only if it is not in $\mathbf{C_A}$, and occurs on the r.h.s only by substituting in for variables not in $\mathbf{C_A}$, $X$ is an ancestor of $A_i$ via a directed path that does not intersect $\mathbf{C_A}$; hence each member of $\mathbf{Parents}(X) \setminus (D(\mathbf{C_A}, \mathbf{A}) \cup \mathbf{C_A})$ is an ancestor of $A_i$ via a directed path that doesn't intersect $\mathbf{C_A}$.

---

$X_2 = 3 X_1 + f_2(\varepsilon_2, X_6)$  $\mathbf{V_1} = \{X_1\}$

$\mathbf{E}_2^1 = \{X_6, \varepsilon_2\}$  $f_2^1(\varepsilon_2, X_6) = f_2(\varepsilon_2, X_6)$  $a_{X_2, X_1}^1 = 3$

Substitute r.h.s of equation for $X_1$ in for $X_1$, in equation for $X_2$

$X_2 = 3 X_1 + f_2(\varepsilon_2, X_6) = 3 (2 L_1 + f_1(\varepsilon_1)) + f_2(\varepsilon_2, X_6) = 6 L_1 + 3 f_1(\varepsilon_1)) + f_2(\varepsilon_2, X_6)$

$\mathbf{V_2} = \{L_1\}$

$\mathbf{E}_2^2 = \{X_6, \varepsilon_2, \varepsilon_1\}$  $f_2^2(X_6, \varepsilon_2, \varepsilon_1) = f_2^1(X_6, \varepsilon_2) + 3 f_2^1(\varepsilon_1)$

$a_{X_2, L_1}^2 = 3 \times 2$  $\mathbf{E_A} = \{X_6, \varepsilon_1, \varepsilon_3\}$

---

**Figure 5:** Illustration of substitutions

After a finite number of substitutions, all of the members of $\mathbf{V_n}$ are members of $\mathbf{C_A}$, and no more substitutions are done. At that stage, by induction,

$$A_i = \sum_{V \in \mathbf{C_A}} a_{i,V} V + f(\mathbf{E}_i)$$

where $\mathbf{E}_i \cap (D(\mathbf{C_A}, \mathbf{A}) \cup \mathbf{C_A}) = \emptyset$, but each member of $\mathbf{E}_i$ is an ancestor of $A_i$ via some path that does not intersect $\mathbf{C_A}$.

Case 2b(ii): $D(\mathbf{C_A}, A_i) \neq \emptyset$. This case now divides into two subcases, $\mathbf{C_B} = \emptyset$ or $\mathbf{C_B} \neq \emptyset$. First consider the case where $\mathbf{C_B} = \emptyset$. Let $\mathbf{E_A}$ be the union of all of the $\mathbf{E}_i$. For each $X \in \mathbf{E}_i$, if there is a trek $T$ between $X$ and $\mathbf{B}$, then it intersects $\mathbf{C_A}$ on the $X$ side, since otherwise $(\mathbf{C_A}; \emptyset)$ does not t-separate $\mathbf{A}$ and $\mathbf{B}$. It follows then that there is a directed path from $\mathbf{C_A}$ to $X$, and $X$ is on a directed path from $\mathbf{C_A}$ to $\mathbf{A}$, contrary to what was proved about each member of $\mathbf{E_A}$. Hence there is no trek between $X$ and $\mathbf{B}$. It follows that $\mathbf{E_A}$ is independent of $\mathbf{B}$, and hence $\mathbf{f}(\mathbf{E_A})$ is independent of $\mathbf{B}$, and $\text{cov}(\mathbf{f}(\mathbf{E_A}), \mathbf{B}) = 0$. Then by Lemma 1, $\text{rank}(\text{cov}(\mathbf{A}, \mathbf{B})) \leq \#\mathbf{C_A}$.

Now suppose $\mathbf{C_B} \neq \emptyset$. Similarly to the case for $\mathbf{A}$, for each $B_i$ in $\mathbf{B}$,

$$B_i = \sum_{V \in \mathbf{C_B}} b_{i,V} V + g_i(\mathbf{E}_i)$$

where each member of $\mathbf{E}_i$ is not in $D(\mathbf{C_B}, \mathbf{B}) \cup \mathbf{C_B}$, but is an ancestor of $B_i$ via some path that does not intersect $\mathbf{C_B}$.

I will now show that for any two members $X$ and $Y$ of $\mathbf{E_A}$ and $\mathbf{E_B}$ respectively, $\text{cov}(X,Y) = 0$. By the construction of $\mathbf{E_A}$ and $\mathbf{E_B}$, there is a directed paths $P_1$ from $X$ to some $A_i$ that does not intersect $\mathbf{C_A}$, and a directed path $P_2$ from $Y$ to some $B_j$ that does not intersect $\mathbf{C_B}$. If $X = Y$, then there is a trek between $\mathbf{A}$ and $\mathbf{B}$ that does not intersect $\mathbf{C_A}$ on the $\mathbf{A}$ side or $\mathbf{C_B}$ on the $\mathbf{B}$ side, contrary to the assumption that $(\mathbf{C_A}; \mathbf{C_B})$ t-separates $\mathbf{A}$ and $\mathbf{B}$. Similarly, if $X \neq Y$, but there is a trek $T$ between $X$ and $Y$, it either intersects $\mathbf{C_A}$ on the $X$ side or $\mathbf{C_B}$ on the $Y$ side, since otherwise $(\mathbf{C_A}; \mathbf{C_B})$ does not t-separate $\mathbf{A}$ and $\mathbf{B}$. But if $T$ intersects $\mathbf{C_A}$ on the $X$ side or $\mathbf{C_B}$ on the $Y$ side, then there is a directed path from $\mathbf{C_A}$ to $X$ or $\mathbf{C_B}$ to $Y$, in which case $X$ is on a directed path from $\mathbf{C_A}$ to $\mathbf{A}$, or $Y$ is on a directed path from $\mathbf{C_B}$ to $\mathbf{B}$, contrary to what was shown about $\mathbf{E_A}$ and $\mathbf{E_B}$. Hence there is no trek between $X$ and $Y$ and $X \neq Y$. It follows that $\mathbf{E_A}$ is independent of $\mathbf{E_B}$, and for any functions $\mathbf{f}$ and $\mathbf{g}$, $\mathbf{f}(\mathbf{E_A})$ is independent of $\mathbf{g}(\mathbf{E_B})$. Hence $\text{cov}(\mathbf{f}(\mathbf{E_A}), \mathbf{g}(\mathbf{E_B})) = 0$. By Lemma 2, $\text{rank}(\text{cov}(\mathbf{A}, \mathbf{B})) \leq \#\mathbf{C_A} + \#\mathbf{C_B}$. Q.E.D.

**Theorem 3:** For all directed graphs $G$, if there does not exist a pair of sets $\mathbf{C'_A}$, $\mathbf{C'_B}$, such that $(\mathbf{C'_A}; \mathbf{C'_B})$ t-separates $\mathbf{A}$ and $\mathbf{B}$ and $\#\mathbf{C'_A} + \#\mathbf{C'_B} \leq r$, then for any $\mathbf{C_A}$, $\mathbf{C_B}$ there is a fixed parameter structural equation model $S$ with path diagram $G$ that is LA below the sets $(\mathbf{C_A}; \mathbf{C_B})$ for $\mathbf{A}$ and $\mathbf{B}$ that entails $\text{rank}(\text{cov}(\mathbf{A},\mathbf{B})) > r$.

**Proof.** $G$ can always be made acyclic by setting the coefficients of edges occurring in cycles to zero. By the Trek Separation Theorem, there is a fixed parameter linear structural equation model $S'$ with path diagram $G$ in which $\text{rank}(\text{cov}(\mathbf{A},\mathbf{B})) > r$. By definition, $S'$ is LA below the sets $\mathbf{C_A}$, $\mathbf{C_B}$ for any $\mathbf{C_A}$, $\mathbf{C_B}$. Q.E.D.

**Acknowledgements:** I wish to thank Rina Foygel, Kelli Talaska, Jan Draisma, Seth Sullivant, and Mathias Drton for substantial help with the proofs at a 2010 AIM workshop on Parameter Identification in Graphical Models.